# Psoriasis Severity Assessment with a Similarity-Clustering Machine Learning Approach Reduces Intra- and Inter-Observer Variation

Arman Garakani[1], Martin Malmstedt-Miller[1,2], Ionela Manole[1,3], Adrian Y. Rössler[1], John R. Zibert[1]

[1] LEO Innovation Lab, Silkegade 8, 1113 Copenhagen, DK
[2] Abdominal Center, Dept. of General Surgery, Herlev Hospital, Copenhagen, Denmark
[3] Dermatology Research Unit, Colentina Clinical Hospital, Bucharest, Romania

**Corresponding author**
Name: Arman Garakani
Address: 15121 Pepper Ln, Saratoga, CA 95070
Email: arman.garakani@darisa.net
Phone: +1 408 201 3371

**Abstract**

**Background:** Psoriasis is a complex disease with many variations in genotype and phenotype. General advancements in medicine have further complicated assessments and treatments for both physicians and dermatologists alike. Even with all our technological progress, we still primarily use the assessment tool Psoriasis Area and Severity Index (PASI) for severity assessments which was developed in the 1970s. In this study, we evaluate a method involving digital images, a comparison web-application and similarity-­clustering, developed to improve the assessment tool in terms of intra- and inter-observer variation. Furthermore, we show that image ranking from pairwise comparisons is applicable to monitoring disease progression over time.

**Materials and methods:** Images of the same lesion area were taken approximately 1 week apart with the patients' own mobile devices. Five dermatologists evaluated the severity of psoriasis in the images by modified-PASI, absolute scoring and a relative pairwise PASI scoring using similarity-clustering and conducted using a web-program displaying two images at a time.

**Results:** mPASI scoring of single images by the same or a different dermatologist showed mPASI ratings of 50% to 80%, respectively. Repeated mPASI comparison using similarity-clustering showed consistent mPASI ratings of >95%. Pearson correlation between absolute scoring and pairwise scoring progression was 0.72.

**Conclusion:** Our similarity-clustering of pairwise relative comparisons by a small number of dermatologists significantly reduces the intra- and inter-observer variation when assessing severity in psoriasis.

**Introduction**

Psoriasis is a chronic, inflammatory and complex skin condition with a substantial burden of disease. Psoriasis has numerous types of clinical manifestations of which chronic plaque-type is the most common, as well as several associated serious comorbidities [1,2]. Treatment and assessment of psoriasis are also complex tasks and are sometimes handled exclusively by dermatologists, but even in that case challenges in assessment of severity and progression occur and physicians as well as dermatologists must rely partly on subjective measures.

Psoriasis Area and Severity Index (PASI), developed by Fredriksson in 1978, still serves as the gold standard for psoriasis severity assessments and is used as a standard in clinical trials [3,4]. Accurate assessments are not only important in clinical trials, but also in the evaluation and choice of treatment in the outpatient clinics and therefore, it is important to have the best tool possible. Clinical studies of psoriasis severity using images, where multiple raters, both human and machine, repeatedly rate the image data according to PASI are widely used to assess treatment efficacy and progression in trials and clinical practice for treatment choices [5]. PASI is widely considered subjective and of poor reproducibility [6]. Difficulty of calculating inter-rater and intra-rater reproducibility remains an obstacle in using PASI [7]. At first glance, values in a variable representing each component do represent presence or absence of disease relatedness and the order is not arbitrary, with 0 = no disease related, 1 = mild, 2 = moderate, 3 = severe and 4 = very severe. Wider close examination has revealed skewed distribution [5,6,7] and inherent subjectivity:

variability in redness and scaling is due to their "relative" nature,
variability in scaliness is due to regional variations and its subjective definition,
variability in thickness is due to inconsistency in the training environment and many others.

Current best practices include PASI assessment training, only using one evaluator per case and use of reference baseline images.

This study evaluates the impact of similarity clustering [8] of all pairwise comparisons in a set of psoriasis images aiming to reduce inter- and intra-rater variability in the context of the PASI components of redness, thickness and scaliness.

**Materials and Methods**
**Image Data**
Images were provided from patients with psoriasis via a digital mobile application (Imagine, Leo Innovation Lab, Copenhagen, Denmark) accepting our terms and conditions to use the pictures for research purposes. An image set for a given patient contained images captured at 5 time points typically 2 weeks apart. 100 image sets were used for calculating repeatability of single scoring and image pair comparison. One image set out of the 100 were used for progression computation.

**Assessment**

Five practicing dermatologists evaluated the severity of psoriasis image sets by modified-PASI (mPASI, limited to scaliness, redness and thickness), absolute scoring and a relative pairwise PASI scoring through similarity clustering facilitated by a web-program (Pairoscope, Leo Innovation Lab, Copenhagen, Denmark).

**Study Design**

For each patient, disease progression, i.e. severity over time, was evaluated twice: using mPASI score calculated from absolute scoring of every image and using similarity-clustering of pairwise relative comparisons of all pairs of images [9]. In all evaluations, patient identification and time of image capture were not revealed.

**Statistical Analysis**

Confusion matrices were used for assessing scoring and agreement among dermatologists. We used multiple ratings by the same dermatologist as multiple independent ratings and focused on inter-rater agreement by treating multiple ratings by the same dermatologist as independent ratings. Progressions of time-ordered image sets were then compared to the corresponding cluster ranks computed from the similarity matrix; a Bradley-Terry model [10]. Pearson Correlation was used for comparing severity progression over time.

**Modified Psoriasis Area Severity Index (mPASI)**

mPASI score was computed by averaging grading for existence of disease related symptoms: scaliness, redness, and thickness (-from 0 to 4 for most severe and integer increments-). Disease progression evaluation for a time-ordered set of images is the corresponding
time-ordered mPASI scores.

**Pairwise Comparisons of mPASI Components**

A Web Interface was used for relative assessment of the pairs of images. The interface displayed two images at a time, allowing zoom and pan of each image independently, 3 sliders with a range of -0.5 to +0.5 set at default position of 0 and a button to submit the comparison. The dermatologist compares the pairs of images as to how similar-/-dissimilar two images are in the context of a disease related symptoms: scaliness, redness, and thickness (-1 to 1.0 in increments of 1/16) mPASI component (see Figure 1). All unique pairings of images are presented to the dermatologist in random order.

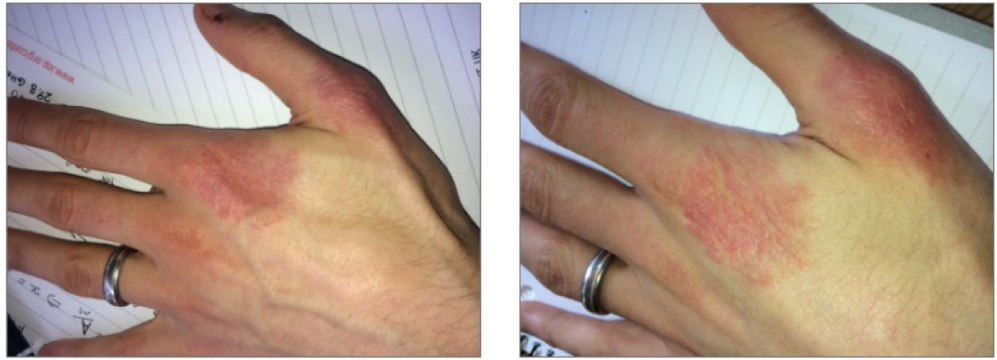

**Disease Progression**

Similarity−clustering and ordering of a collection of items is equivalent to ranking items in a collection from pairwise probabilities [12]. We define disease progression as a time series of normalized severity (0.0 - 1.0). Normalized severity has the following properties:

P(t) = 0.0 implies disease progression at time t is a minimum in the sequence, and

P(t) = 1.0 implies disease progression at time t is a maximum in the sequence

**Pairwise Scoring Algorithm**

Input to the algorithm is a set of images captured overtime [n] = {1,...,n}. Pairwise comparison, $M_{ij} \in [0,1]$, sets the probabilities that image *i* is higher than image *j* in the current context and that *j* beats *i* with the probability of 1 - $M_{ij}$. A similarity matrix is formed from all pairwise comparisons. For each image, the ranking score is the probability that an image has higher progression than any other image picked uniformly random from other images. It is simply calculated by averaging all of an image's pairwise comparisons.

$$\text{Ranking Score} = 1 / (n-1) \sum M_{ij} \text{ for all } j \neq i$$

Disease Progression for the image captured at time t, P(t) is the Ranking Score for image taken at time t.

**Numeric Example**

The similarity matrix from pairwise comparisons:

    [ 0.5  0.7  0.6  0.8  0.7]
    [ 0.3  0.5  0.8  0.7  0.6]
    [ 0.4  0.2  0.5  0.8  0.7]
    [ 0.2  0.3  0.2  0.5  0.4]
    [ 0.3  0.4  0.3  0.6  0.5]]

Ranking Score：   [ 1.7  2.1  2.4  3.4  2.9]
Ranks:   [0 1 2 4 3]

**Process for one dermatologist using the Web Interface:**
For each context in contexts ( **redness, scaliness or thickness** )
    Compare all pairs (i , j ) were i is not equal to j

**Results and Discussion**
Redness and thickness compare reasonably well in localization of most severe time points. 100 image sets of on average of 5 images captured over time were used to compute repeatability of single images by repeated scoring by 5 dermatologists. The same image sets were used to compute repeatability of comparison scoring. Repeated mPASI scoring of single images by the same or different dermatologist showed consistent mPASI ratings of 50% to 80%, respectively. Repeated mPASI comparison using the similarity--clustering program showed consistent mPASI ratings of >95%. Confusion matrices of single scoring and pairwise scoring is shown in Figure 2. Confusion matrices of Agreement among dermatologists for both single and pairwise scoring is shown in the top row in Figure 3. It demonstrates the finer grain classification of pairwise scoring. We found Pearson Correlation to overestimate the degree of agreement among raters using the single scoring method and propose Total Deviation Index [5–7] (see Figure 3).

A single image set was used to compare severity progression generated via pairwise protocol with single scoring protocol. In the single scoring protocol of an image set, modified PASI score is generated from single image scoring of every image in the set by a dermatologist. Pearson correlation between single scoring protocol and pairwise scoring protocol computed progression trends was 0.72.

**Conclusions**

Assessment of disease severity remains a multifactorial discipline involving genetics, immunology, quality of life, amongst other factors, but we might be able to assist the assessment with a tool like pairwise comparison.

According to the present study, a simple web application can significantly improve severity assessments of images of psoriasis captured with a cellphone camera by the patients. Our approach is practical for measuring disease progression from time lapse images. In many larger and practical situations the true pairwise comparisons can-not be practically measured, however when a subset of all comparisons is passively and noisily observed, a good estimate can be calculated-[11]. If the image set needs to be partitioned into a number of disjoint sets, the Acting Ranking [12] can be used.

Our similarity clustering program, also known as ranking from pairwise comparisons, significantly reduces the intra- and inter-observer variation when assessing severity in psoriasis, which indicates that the application in clinical trials and practice could be useful. Emerging application of machine learning in dermatology depends critically on high quality labeling of the training data, i.e. image assessment by multiple dermatologists. Our study demonstrates that pairwise comparisons may be better suited to generating such data than single scoring methods laying the foundation for application of computer vision algorithms to automate large scale severity assessments.

**Data Availability**

The data used to support the findings of this study are available from the corresponding author upon request.

**Conflicts of Interest**

The authors are employees or former employees of LEO Innovation Lab, an independently established innovation unit of LEO Pharma A/S. The project was funded by LEO Innovation Lab.

**Funding Statement**

LEO Innovation Lab has funded this study.

**Acknowledgments**

We would like to thank the dermatologists who have participated in making severity assessments and also the users of and the team behind the app Imagine.

# References

1. Takeshita J, Grewal S, Langan SM, Mehta NN, Ogdie A, Van Voorhees AS, et al. Psoriasis and comorbid diseases: Epidemiology. J Am Acad Dermatol [Internet]. 2017 Mar;76(3):377–90. Available from: http://dx.doi.org/10.1016/j.jaad.2016.07.064

2. Schmitt J, Wozel G. The Psoriasis Area and Severity Index Is the Adequate Criterion to Define Severity in Chronic Plaque-Type Psoriasis. Dermatology [Internet]. 2005 Apr 11;210(3):194–9. Available from: https://www.karger.com/Article/FullText/83509

3. Fredriksson T, Pettersson U. Severe psoriasis--oral therapy with a new retinoid. Dermatology [Internet]. 1978;157(4):238–44. Available from: https://www.karger.com/Article/Abstract/250839

4. Feldman SR, Krueger GG. Psoriasis assessment tools in clinical trials. Ann Rheum Dis [Internet]. 2005 Mar 1 [cited 2018 Jul 3];64(suppl 2):ii65–8. Available from: https://ard.bmj.com/content/64/suppl_2/ii65

5. Ribas J, Cunha M da GS, Schettini APM, Ribas CB da R. Agreement between dermatological diagnoses made by live examination compared to analysis of digital images. An Bras Dermatol [Internet]. 2010 Jul;85(4):441–7. Available from: https://www.ncbi.nlm.nih.gov/pubmed/20944903

6. Youn SW, Choi CW, Kim BR, Chae JB. Reduction of Inter-Rater and Intra-Rater Variability in Psoriasis Area and Severity Index Assessment by Photographic Training. Ann Dermatol [Internet]. 2015 Oct;27(5):557–62. Available from: http://dx.doi.org/10.5021/ad.2015.27.5.557

7. Gourraud P-A, Le Gall C, Puzenat E, Aubin F, Ortonne J-P, Paul CF. Why statistics matter: limited inter-rater agreement prevents using the psoriasis area and severity index as a unique determinant of therapeutic decision in psoriasis. J Invest Dermatol [Internet]. 2012 Sep;132(9):2171–5. Available from: http://dx.doi.org/10.1038/jid.2012.124

8. Balcan M-F, Blum A, Vempala S. Clustering via similarity functions: theoretical foundations and algorithms. In: 40th ACM Symposium on Theory of Computing Conference [Internet]. 2008. p. 17–20. Available from: http://www.cs.cmu.edu/afs/cs.cmu.edu/Web/People/avrim/Papers/BBVclustering_full.pdf

9. Ammar A, Shah D. Ranking: Compare, don't score. In: 2011 49th Annual Allerton Conference on Communication, Control, and Computing (Allerton) [Internet]. 2011. p. 776–83. Available from: http://dx.doi.org/10.1109/Allerton.2011.6120246

10. Bradley RA, Terry ME. Rank Analysis of Incomplete Block Designs: I. The Method of Paired Comparisons. Biometrika [Internet]. 1952;39(3/4):324–45. Available from: http://www.jstor.org/stable/2334029

11. Wauthier F, Jordan M, Jojic N. Efficient Ranking from Pairwise Comparisons. In: International Conference on Machine Learning [Internet]. 2013 [cited 2018 Jul 3]. p. 109–17. Available from: http://proceedings.mlr.press/v28/wauthier13.html

12. Heckel, Reinhard; Shah, Nihar B.; Ramchandran, Kannan; Wainwright, Martin J. Active ranking from pairwise comparisons and when parametric assumptions do not help. Ann. Statist. 47 (2019), no. 6, 3099--3126. doi:10.1214/18-AOS1772. https://projecteuclid.org/euclid.aos/1572487385

Figure 2: Confusion matrix among dermatologists for standard and pairwise scoring from dermatologists(A-E) of redness, thickness and scaling over 5 time points (hours from first image).

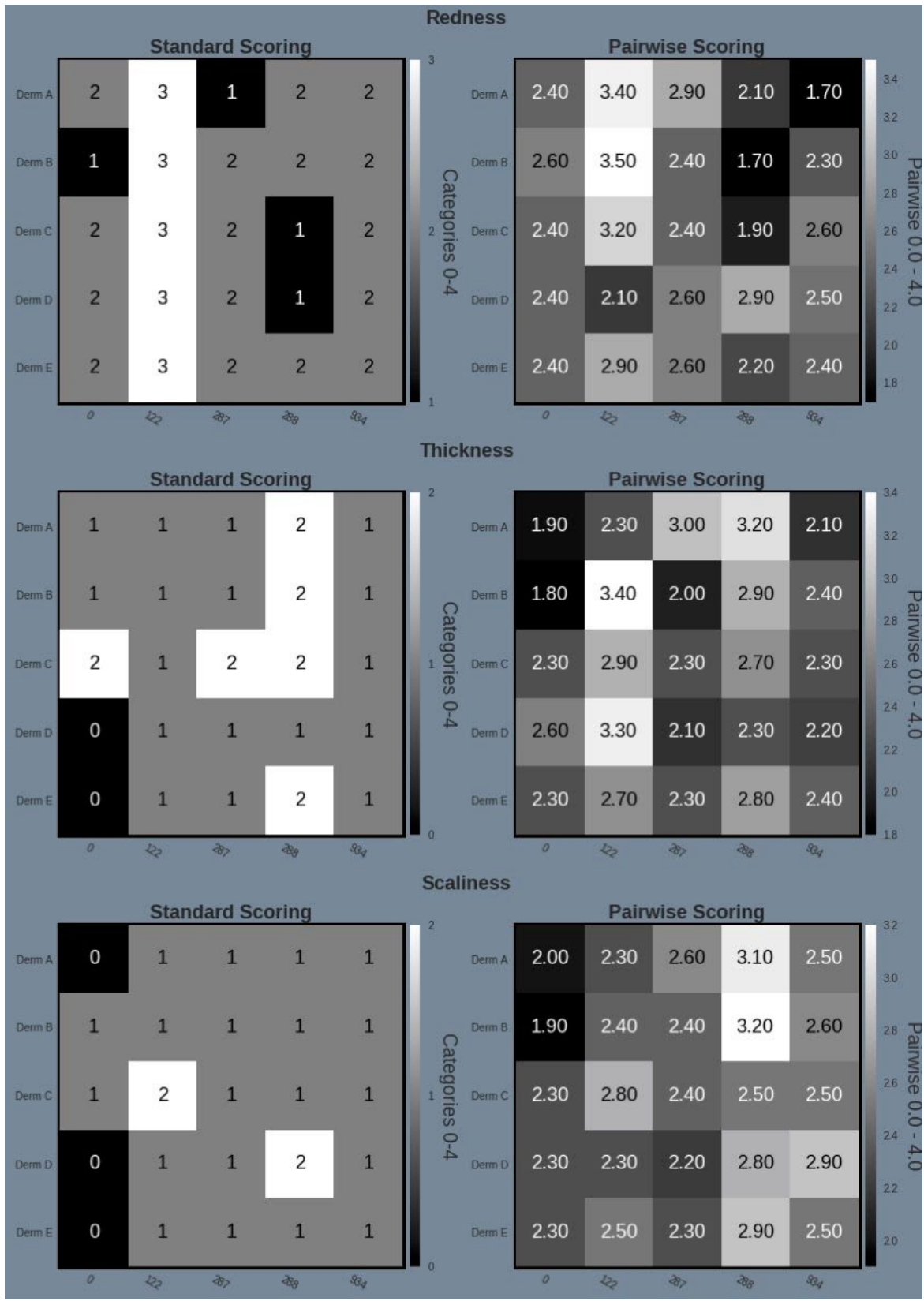

Figure 3: Agreement Among Raters and Over-Estimation of Agreement

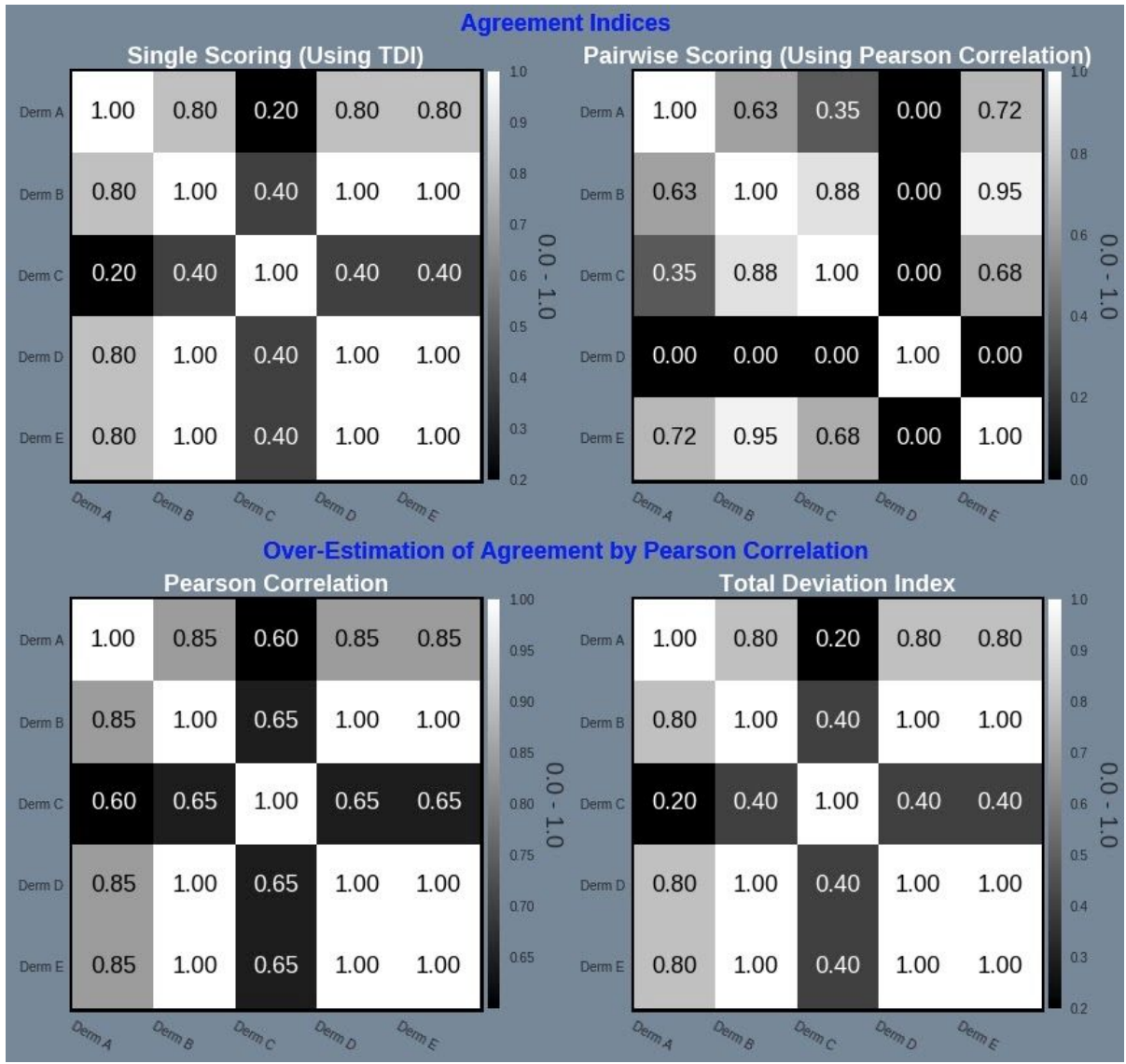